\documentclass[15pt]{article}
\usepackage{spconf}
\usepackage{citesort}
\usepackage[fleqn]{amsmath}
\usepackage{ascmac}
\usepackage[dvips]{graphicx}
\usepackage{psfrag}
\usepackage{amsmath}
\usepackage{amsbsy}
\usepackage{amssymb}
\usepackage{latexsym}
\usepackage{subfigure}
\usepackage{comment}

\graphicspath{{./images/}}

\newtheorem{assumption}{Assumption}

\newtheorem{theorem}{Theorem}

\newcommand{\migip}{\hspace*{\fill} $\Box $}

\newcommand{\euclidspace}{{\mathcal{H}}}
\newcommand{\signal}[1]{{\boldsymbol{#1}}}
\newcommand{\Natural}{{\mathbb N}}
\newcommand{\norm}[1]{\left\|#1\right\|}
\newcommand{\abs}[1]{\left|#1\right|}

\newcommand{\real}{\ensuremath{{\mathrm{I\!R}}}}

\newcommand{\innerprod}[2]{\left\langle{#1},{#2}\right\rangle}

\newcommand{\refeq}[1]{(\ref{#1})}

\newcommand{\sinq}[1]{`#1'}

\def\baselinestretch{.8}

\title{A STOCHASTIC BEHAVIOR ANALYSIS OF STOCHASTIC RESTRICTED-GRADIENT
DESCENT ALGORITHM IN REPRODUCING KERNEL HILBERT SPACES}
\name{
Masa-aki Takizawa$^\dagger$, Masahiro Yukawa$^\dagger$\sthanks{This work was partially supported by
JSPS Grants-in-Aid (24760292).},
and C\'edric Richard$^{\ddagger}$}
\address{$^\dagger$ Department of Electronics and Electrical Engineering, Keio University, Japan\\
$^\ddagger$ Universit\'e de Nice Sophia-Antipolis, CNRS, France}
\begin{document}
\ninept 
\maketitle
\begin{abstract}
\vspace{-.2cm}
This paper presents a stochastic behavior analysis of 
a kernel-based stochastic restricted-gradient descent method.
The restricted gradient gives a steepest ascent direction within
the so-called dictionary subspace.
The analysis provides the transient and steady state performance
in the mean squared error criterion.
It also includes stability conditions in the mean and mean-square sense.
The present study is based on the analysis of the kernel normalized least
 mean square (KNLMS) algorithm initially proposed by Chen {\it et al.}
Simulation results validate the analysis.
\end{abstract}



%

\vspace{-.2cm}
\section{Introduction}
\vspace{-.2cm}
Kernel adaptive filtering \cite{liu_book10} is an attractive approach
for nonlinear estimation problems based on the theory of
reproducing kernel Hilbert space (RKHS), and 
a number of kernel adaptive filtering algorithms 
have been proposed
\cite{kivinen04,engel04,richard09,slavakis09,yukawa_tsp12,chen_TNNLS12,steven}.
The existing kernel adaptive filtering algorithms
are classified into two general categories 
according to the space in which optimization is performed \cite{yukawa_tsp12}:
(i) the RKHS approach
(e.g., \cite{kivinen04,slavakis09,chen_TNNLS12}) and 
(ii) the parameter-space approach
(e.g., \cite{richard09,yukawa_tsp12,gao2014l1KLMS}).
The kernel normalized least  mean square (KNLMS) algorithm
is a representative example of the parameter-space approach
and its stochastic behavior analyses have been presented in
\cite{analysisKNLMS,richard2012closed,chen2014}.
The analyses have clarified the transient and steady-state performance
in the mean squared error (MSE).
A stochastic restricted-gradient descent algorithm studied in
the present work is an RKHS counterpart of the KNLMS algorithm.
We call it the natural kernel least mean square (Natural KLMS) algorithm
to distinguish it from the KLMS algorithm proposed in \cite{liu_TSP08}.
A primitive question is whether it is possible to give the same analyses
as in \cite{analysisKNLMS,richard2012closed,chen2014}
for the stochastic restricted-gradient descent algorithm.
If this is possible, it will provide a theoretical basis to
compare the performances of KNLMS and Natural KLMS.
This will eventually give a new insight into 
the relationship between the two classes of kernel adaptive filtering algorithms.

To clarify the orientation of the Natural KLMS algorithm
in the kernel adaptive filtering researches,
let us give a short note on the RKHS approach.
Dictionary sparsification is a common issue of kernel adaptive
filtering \cite{platt,engel04,richard09,liu_book10}.
The KLMS algorithm \cite{liu_TSP08}
updates the filter only when the current input datum is added into the
dictionary and this would cause severe performance degradations.
A systematic scheme which eliminates such a limitation has been proposed
in \cite{hypass} under the name of {\it hyperplane projection along
affine subspace (HYPASS)}.
The HYPASS algorithm updates the filter using the projection onto the
zero-instantaneous-error hyperplane along the so-called 
{\it dictionary subspace} $\mathcal{M}$,
the subspace spanned by the dictionary elements.
This is achieved by projecting the gradient direction onto $\mathcal{M}$.
In a nutshell, HYPASS is the NLMS algorithm operated in the dictionary subspace $\mathcal{M}$.
Natural KLMS is actually an LMS counterpart of HYPASS and we consider
this LMS-based algorithm to make the analysis feasible.
In \cite{KLMSanalysis} and \cite{chen_TNNLS12}, 
the mean square convergence analysis and the theoretical steady-state
MSE have been presented for the KLMS and Quantized KLMS algorithms, respectively.
However, transient performance analyses have not yet been reported
due to the difficulty in treating the growing number of dictionary elements.
     
In this paper, we present a stochastic behavior analysis of 
the Natural KLMS algorithm with a Gaussian kernel under i.i.d.~random inputs
based on the framework presented in \cite{chen2014}.
Natural KLMS is derived by using the {\it restricted gradient}
which gives a steepest ascent direction
within the dictionary subspace $\mathcal{M}$.
The analysis provides theoretical MSEs during the transient phase as well as at the steady-state.
We also derive stability conditions in the mean and mean-square sense.
The key ingredients for the analysis are the restricted gradient
and the isomorphism between the dictionary subspace $\mathcal{M}$ and
a Euclidean space;
these were also the key when the first and second authors
developed a sparse version of HYPASS in \cite{takizawa2014}.
The validity of the analysis is illustrated by simulations.

\vspace{-.2cm}
\section{Preliminaries}
\vspace{-.2cm}
We address an adaptive estimation problem of a nonlinear system $\psi$
with sequentially arriving input signals $\signal{u}\in
\mathcal{U}\subset \real^L$, and its noisy output
$d:=\psi(\signal{u})+\nu \in\real$, where 
$\signal{u}$ is assumed an i.i.d.~random vector and $\nu$ is a zero-mean
additive noise uncorrelated with any other signals.
The function $\psi$ is modeled as an element of the RKHS $\mathcal H$
associated with a Gaussian kernel
$  \kappa(\signal{x},\signal{y}):=\exp\left(-\dfrac{\norm{\signal{x}-\signal{y}}^2}{2\sigma^2}\right),\ \signal{x},\signal{y}\in \mathcal{U}$,
where $\sigma>0$ is the kernel parameter.
We denote by $\innerprod{\cdot}{\cdot}$ and $\norm{\cdot}$ the canonical
inner product and the norm defined in $\real^L$, respectively, and
$\innerprod{\cdot}{\cdot}_{\mathcal{H}}$ and $\norm{\cdot}_\mathcal{H}$
those in $\mathcal{H}$.
A kernel adaptive filter is given as a finite order filter:
\begin{equation}
\varphi_n := \sum^{}_{j\in\mathcal{J}}\alpha^{(n)}_{j}\kappa(\cdot,\signal{u}_j),\ n\in\Natural,\label{eq:kaf}
\end{equation}
where $\alpha^{(n)}_{j}\in \real$ are the filter coefficients and $\mathcal{J}:=\{j_{1},j_{2},\cdots,j_{r}\}$ indicates the dictionary
$\{\kappa(\cdot,\signal{u}_j)\}_{j\in\mathcal{J}}$; $n$ is the time index.
Without loss of generality, we assume that the dictionary is a linearly
independent set so that it spans an $r$ dimensional subspace
\begin{equation}
\mathcal{M} := {\rm
 span}\{\kappa(\cdot,\signal{u}_j)\}_{j\in\mathcal{J}}\subset\mathcal{H},\label{eq:dic_subspace}
\end{equation}
which is called the {\it  dictionary subspace}.
Although the dictionary is updated typically during the learning process,  
we assume that the dictionary is fixed to make the analysis tractable.

The instantaneous error at time instant $n$ is defined as
$e_n := d_n-\innerprod{\varphi}{\kappa(\cdot,\signal{u}_n)}_{\mathcal{H}}=d_n-\innerprod{\signal{\alpha}}{\signal{\kappa}_n}$,
where
$\signal{\kappa}_n=[\kappa(\signal{u}_n,\signal{u}_{j_1}),$
$\kappa(\signal{u}_n,\signal{u}_{j_2}),\cdots,\kappa(\signal{u}_n,\signal{u}_{j_r})]^{\sf
T}$ is the vector of the kernelized input and
$\signal{\alpha}=[\alpha_{j_1}, \alpha_{j_2},\cdots,\alpha_{j_r}]^{\sf
T}$ is the coefficient vector. 
The MSE cost function, with respect to the coefficient vector $\signal{\alpha}$,
 is given by
\begin{equation}
J(\signal{\alpha}):= E(e^2_n(\signal{\alpha})) =
E(d^2_n) +\signal{\alpha}^{\sf T} \signal{R}_{\kappa} \signal{\alpha} - 2\signal{p}^{\sf T}\signal{\alpha}
\label{eq:MSE_para},
\end{equation}
where $\signal{R}_{\kappa} := E(\signal{\kappa}_n \signal{\kappa}^{\sf T}_n)$ is the
autocorrelation matrix of the
kernelized input $\signal{\kappa}_n$ and $\signal{p} := E(d_n\signal{\kappa}_n)$ is the
cross-correlation vector between $\signal{\kappa}_n$ and $d_n$.
With the optimization in RKHS in mind,
the MSE, with respect to $\varphi$, is given by:
\begin{align}
J(\varphi) := E(e^2_n(\varphi))= &E({d^2_n}) +
E(\innerprod{\varphi}{\kappa(\cdot,\signal{u}_n)}^2_{\mathcal H})  \nonumber \\ &- 2E(d_n\innerprod{\varphi}{\kappa(\cdot,\signal{u}_n)}_{\mathcal H})\label{eq:MSE_rkhs}.
\end{align}

While the KNLMS algorithm optimizes $J(\signal{\alpha})$ in the Euclidean space $\real^L$,
the Natural KLMS algorithm presented in the following section 
optimizes $J(\varphi)$ in the RKHS $\euclidspace$ under the restriction
to the dictionary subspace $\mathcal{M}$, or in short,
it optimizes $J(\varphi)$ in $\mathcal{M}$.
Referring to \cite{kivinen04},
the stochastic gradient descent method for $J(\varphi)$ in
$\mathcal{H}$ updates the filter $\varphi_n$ along the \sinq{line}
(one dimensional subspace) spanned by the singleton
$\{\kappa(\cdot,\signal{u}_n)\}$.
This implies that the filter is updated
only when $\kappa(\cdot,\signal{u}_n)$ is added
into the dictionary, because otherwise $\varphi_n +
\alpha\kappa(\cdot,\signal{u}_n)\not\in\mathcal{M}$
for any $\alpha\neq 0$.
We thus present the restricted gradient,
which was initially introduced in \cite{takizawa2014},
and derive the Natural KLMS algorithm in the following section.

\vspace{-.1cm}
\section{The Natural KLMS algorithm}
\vspace{-.1cm}
The ordinary gradient of $J(\signal{\alpha})$ in $\real^r$ is given by
 $\nabla J(\signal{\alpha})= 2(\signal{R}_{\kappa}\signal{\alpha}
-\signal{p})$.
Given any positive definite matrix $\signal{Q}$,
 $\innerprod{\signal{x}}{\signal{y}}_{\signal{Q}}:=\signal{x}^{\sf
 T}\signal{Q}\signal{y}$ and
 $\norm{\signal{x}}_{\signal{Q}}:=\sqrt{\signal{x}^{\sf
 T}\signal{Q}\signal{x}}$ define an inner product and its induced norm,
 respectively.
The $\signal{G}$-gradient of \refeq{eq:MSE_para} with the inner product
 $\innerprod{\cdot}{\cdot}_\signal{G}$ is defined as \cite{takizawa2014}
\begin{equation}
  \label{eq:Ginner}
  \nabla_{\signal{G}}J(\signal{\alpha}) := \signal{G}^{-1}\nabla J(\signal{\alpha}),
\end{equation}
where $[\signal{G}]_{\ell,m}=\kappa(\signal{u}_{j_\ell},\signal{u}_{j_m})$ for $1\leq \ell,m\leq r$ is the Gram matrix.\footnote{The Gram matrix $\signal{G}$
is ensured to be positive definite due to the assumption
that the elements of the dictionary are linearly independent.
The definition of the $\signal{G}$-gradient is validated by observing that
$\innerprod{\signal{\beta}-\signal{\alpha}}{\nabla_{\signal{G}}J(\signal{\alpha})}_{\signal{G}}
+ J(\signal{\alpha}) = 
\innerprod{\signal{\beta}-\signal{\alpha}}{\nabla J(\signal{\alpha})}
+ J(\signal{\alpha})\leq J(\signal{\beta})
$ for any $\signal{\beta}\in\real^L$.
}

\begin{figure}[t]
  \centering
\psfrag{phi}{\hspace{-.1cm}$\varphi=\sum_{j\in\mathcal{J}}\alpha_j\kappa(\cdot,\signal{u}_j)$}
\psfrag{alpha}{$\signal{\alpha}=[\alpha_{j_1},\alpha_{j_2},\cdots,\alpha_{j_r}]^{\sf T}$}
\psfrag{nabla}{\hspace{.5cm}$\nabla_{|\mathcal{M}} J(\varphi)$}
\psfrag{M}{$\mathcal{M}$}
\psfrag{R}{$\real^r$}
\psfrag{rest}{$\nabla_{\signal{G}} J (\signal{\alpha}):=\signal{G}^{-1}\nabla J(\signal{\alpha})$}
\includegraphics[width=8.5cm]{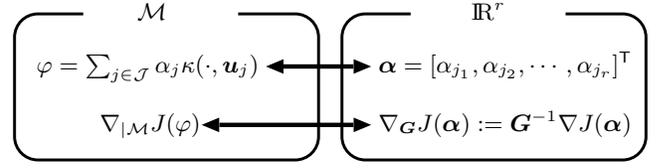}
\caption{The isomorphism between $\real^r$ and $\mathcal{M}$ and
 the restricted gradient.}
\label{fig:restrict}
\end{figure}

The functional Hilbert space
$\left(\mathcal{M},\innerprod{\cdot}{\cdot}_{\mathcal{H}}\right)$
of dimension $r$ is
isomorphic to the Hilbert space
$\left(\real^{r},\innerprod{\cdot}{\cdot}_{\signal{G}}\right)$
under the correspondence (see Fig.~\ref{fig:restrict})
\begin{equation}
 \mathcal{M}\ni \varphi :=
\sum^{}_{j\in\mathcal{J}}\alpha_{j}\kappa(\cdot,\signal{u}_j)
 \longleftrightarrow
 \signal{\alpha}:=
[\alpha_{j_1},\cdots,\alpha_{j_r}]^{\sf T}\in\real^{r}.
\label{eq:isomorphism}
\end{equation}
Note here that the isomorphism as Hilbert spaces includes,
in addition to the one-to-one correspondence between the elements,
the preservation of the inner product; i.e., 
$\innerprod{\varphi_1}{\varphi_2}_{\euclidspace}
=\innerprod{\signal{\alpha}_1}{\signal{\alpha}_2}_{\signal{G}}$
for any $\varphi_1\longleftrightarrow \signal{\alpha}_1$ and
$\varphi_2\longleftrightarrow \signal{\alpha}_2$.
Under the correspondence in \refeq{eq:isomorphism},
the restricted gradient $\nabla_{|\mathcal{M}} J(\varphi)$
is defined, through the $\signal{G}$-gradient in $\real^L$, as follows \cite{takizawa2014}:
\begin{equation}
 \nabla_{|\mathcal{M}} J(\varphi)\longleftrightarrow  \nabla_{\signal{G}} J (\signal{\alpha})=
\signal{G}^{-1}\nabla J(\signal{\alpha}).
\end{equation}
The {\it restricted gradient} $\nabla_{|\mathcal{M}} J(\varphi)$
gives the steepest ascent direction,
within the dictionary subspace $\mathcal{M}$, 
of the tangent plane of the functional \refeq{eq:MSE_rkhs} at the point
$\varphi$. See the derivation of the restricted gradient in \cite{takizawa2014}.
An instantaneous approximation of the restricted gradient
$\nabla_{|\mathcal{M}} J(\varphi_n)\longleftrightarrow
\nabla_{\signal{G}} J (\signal{\alpha}_n)$,
where $\signal{\alpha}_n:=
[\alpha_{j_1}^{(n)},\alpha_{j_2}^{(n)},\cdots,\alpha_{j_r}^{(n)}]^{\sf T}\in\real^{r}$
 is given by 
$\tilde{\nabla}_{|\mathcal{M}} J(\varphi_n)\longleftrightarrow
\tilde{\nabla}_{\signal{G}} J(\signal{\alpha}_n):=
\signal{G}^{-1}\tilde{\nabla} J(\signal{\alpha}_n):=
2\signal{G}^{-1}(\signal{\kappa}_n\signal{\kappa}_n^{\sf
T}\signal{\alpha}_n
-  d_n\signal{\kappa}_n)= -2e_n \signal{G}^{-1}\signal{\kappa}_n$.
Hence, for the initial vector $\signal{\alpha}_0:=\signal{0}$,
 the stochastic restricted-gradient descent method,
which we call the {\it Natural KLMS algorithm}, is given by
\begin{equation}
\signal{\alpha}_{n+1}:= \signal{\alpha}_n - \frac{\eta}{2}
\tilde{\nabla}_{\signal{G}} J(\signal{\alpha}_n)
=\signal{\alpha}_{n} +\eta e_n\signal{G}^{-1}\signal{\kappa}_n,
~~~n\in\Natural,
\label{eq:hypass} 
\end{equation}
where $\eta>0$ is the step size.
The Natural KLMS algorithm \refeq{eq:hypass} requires $r^2$ complexity
for each time update, and this would make a significant impact on the
overall complexity of the algorithm. 
In \cite{hypass,takizawa_tsp_2014}, a simple selective-updating idea for complexity
reduction without serious performance degradations has been presented;
it will be shown in Section
\ref{sec:numerical} that the selective-updating works well.
\vspace{-.1cm}  
\section{Performance analysis}
\vspace{-.1cm}
\subsection{Key idea and assumption}
\vspace{-.1cm}
We derive a theoretical MSE and stability conditions for the Natural KLMS algorithm given by \refeq{eq:hypass} with Gaussian kernel, given the dictionary $\{\kappa(\cdot,\signal{u}_{j})\}_{j\in\mathcal{J}}$.
Left-multiplying both-sides of \refeq{eq:hypass} by the square root
$\signal{G}^{\frac{1}{2}}$ of $\signal{G}$ yields\footnote{For any positive semi-definite matrix
$\signal{Q}$, there exists a unique square root
$\signal{Q}^{\frac{1}{2}}$ satisfying
$\signal{Q}=\signal{Q}^{\frac{1}{2}}\signal{Q}^{\frac{1}{2}}$.}
\begin{equation}
\tilde{\signal{\alpha}}_{n+1} = \tilde{\signal{\alpha}}_n + \eta e_n\tilde{\signal{\kappa}}_n,\label{eq:hypass2}
\end{equation}
 where $\tilde{\signal{\kappa}}_n=\signal{G}^{-\frac{1}{2}}\signal{\kappa}_n$, $\tilde{\signal{\alpha}}_n = \signal{G}^{\frac{1}{2}}\signal{\alpha}_n$.
The cost function $J(\signal{\alpha})$ in \refeq{eq:MSE_para} can be rewritten by 
\begin{equation}
\left(J(\signal{\alpha})=\right)\ \tilde{J}(\tilde{\signal{\alpha}})=E(d_n^2)+\tilde{\signal{\alpha}}^{\sf{T}}\tilde{\signal{R}}_{\kappa}\tilde{\signal{\alpha}}-2\tilde{\signal{p}}^{\sf{T}}\tilde{\signal{\alpha}},
\label{eq:costrkhs}
\end{equation}
as a function of $\tilde{\signal{\alpha}}:=\signal{G}^{\frac{1}{2}}\signal{\alpha}$, and \refeq{eq:hypass2} can be regarded as a stochastic gradient descent method for this cost function $\tilde{J}(\tilde{\signal{\alpha}})$.
Here
\begin{equation}
\tilde{\signal{R}}_{\kappa}:=E(\tilde{\signal{\kappa}}_n\tilde{\signal{\kappa}}_n^{\sf{T}})=\signal{G}^{-\frac{1}{2}}\signal{R}_{\kappa}\signal{G}^{-\frac{1}{2}},\label{rhat}\end{equation}
and
\begin{equation}
  \tilde{\signal{p}}:=E(d_n\tilde{\signal{\kappa}}_n)=\signal{G}^{-\frac{1}{2}}\signal{p},
\end{equation}
are the autocorrelation matrix and the cross-correlation vector for the modified vector $\tilde{\signal{\kappa}}_n$, respectively. 

As $\tilde{\signal{R}}_{\kappa}$ is positive definite \cite{analysisKNLMS}, the optimum weight vector is given by
\begin{equation}
  \label{eq:optalpha}
\tilde{\signal{\alpha}}^* :=\tilde{\signal{R}}_{\kappa}^{-1}\tilde{\signal{p}}, 
\end{equation}  
and with $\tilde{\signal{\alpha}^*}$, we define the weight error vector
\begin{equation} 
\tilde{\signal{v}}_n:=\tilde{\signal{\alpha}}_n-\tilde{\signal{\alpha}}^*.
\end{equation}
In the present analysis, $\tilde{\signal{\kappa}}_n\tilde{\signal{\kappa}}^{\sf T}_n$ needs to be independent of $\tilde{\signal{v}}_n$, which is guaranteed by making the following conditioned modified independence assumption (CMIA) \cite{chen2014}.
\begin{assumption}
 $\signal{\kappa}_n\signal{\kappa}^{\sf T}_n$ is independent of $\signal{v}_n(=\signal{G}^{-\frac{1}{2}}\tilde{\signal{v}}_n)$.
\end{assumption}

\vspace{-.25cm}
\subsection{Mean weight error analysis}
\vspace{-.2cm}
The estimation error can be expressed by 
\begin{equation}
e_n = d_n -\tilde{\signal{\kappa}}_n^{\sf T}\tilde{\signal{v}}_n-\tilde{\signal{\kappa}}_n^{\sf T}\tilde{\signal{\alpha}}^* \label{eq:error}.
\end{equation}
Substituting \refeq{eq:error}
 to \refeq{eq:hypass2}, we obtain the recursive expression for $\tilde{\signal{v}}_n$: 
\begin{equation}
\tilde{\signal{v}}_{n+1} = \tilde{\signal{v}}_{n}  + \eta d_n\tilde{\signal{\kappa}}_n-\eta\tilde{\signal{\kappa}}^{\sf T}_n\tilde{\signal{v}}_n\tilde{\signal{\kappa}}_n-\eta\tilde{\signal{\kappa}}^{\sf T}_n\tilde{\signal{\alpha}}^{*}\tilde{\signal{\kappa}}_n.
\end{equation}
Using CMIA, we obtain the mean weight error model
\begin{equation}
E(\tilde{\signal{v}}_{n+1}) = (\signal{I}_r-\eta\tilde{\signal{R}}_{\kappa})E(\tilde{\signal{v}}_n),\label{eq:recv}
\end{equation}
where $\signal{I}_{r}$ denotes the $r\times r$ identity matrix for any positive integer $r$. 
Let the input $\signal{u}_n$ be a random vector following a Gaussian
distribution with zero mean and the covariance matrix
$\signal{R}_{u}:=E(\signal{u}_n\signal{u}_n^{\sf T})$.
 Then, the $(\ell,m)$ component ($1\leq \ell,m\leq r$) of
the autocorrelation matrix $\signal{R}_{\kappa}$ of $\signal{\kappa}_n$ is given by \cite{chen2014}:\vspace{-.25cm}
 \begin{align}
\centering 
   \label{eq:R}
   &[\signal{R}_{\kappa}]_{\ell,m}=|\signal{I}_L+\frac{2}{\sigma^2}\signal{R}_{u}|^{-\frac{1}{2}}
  \nonumber \\
  &\exp\left[-\frac{1}{4\sigma^2}\left(2\norm{\bar{\signal{u}}_{\ell
  m}}^{(2)}-\norm{\bar{\signal{u}}_{\ell
  m}}^2_{\left(\signal{I}_L+\frac{\sigma^2}{2}\signal{R}^{-1}_{u}\right)^{-1}}\right)\right],\nonumber
 \end{align}
where $\bar{\signal{u}}_{\ell m}=\signal{u}_{j_\ell}+\signal{u}_{j_m}$,
 $\norm{\bar{\signal{u}}_{\ell
m}}^{(2)}=\norm{\signal{u}_{j_\ell}}^2+\norm{\signal{u}_{j_m}}^2$,
and $\abs{\cdot}$ stands for determinant.

From the recursion in \refeq{eq:recv}, we obtain the mean stability
condition of the Natural KLMS algorithm as follows.
\vspace{-.15cm}
\begin{theorem}[Stability in the mean]
Assume CMIA holds. Then, for any initial condition, given dictionary
 $\{\kappa(\cdot,\signal{u}_j)\}_{j \in \mathcal{J}}$, the Natural KLMS
 algorithm asymptotically converges in the mean if the step size is
 chosen to satisfy \vspace{-.15cm}
\begin{equation}
0<\eta<\frac{2}{\lambda_{{\rm max}}(\tilde{\signal{R}}_{\kappa})},
\label{eq:theorem1}
\end{equation}\vspace{-.2cm}
where $\lambda_{\rm max}(\cdot)$ denotes the maximum eigenvalue of the matrix.
\end{theorem}
{\it Proof:} It is clear from the well-known mean stability
results (see, e.g., \cite{sayed}).

\vspace{-.15cm} 
\subsection{Mean-square error analysis}
\vspace{-.15cm}
Squaring \refeq{eq:error} and 
taking its expectation under CMIA, the MSE
\refeq{eq:costrkhs} of Natural KLMS can be rewritten as
\begin{equation}
\label{eq:mse}
\tilde{J}(\tilde{\signal{\alpha}}_n)
=J_{{\rm min}}+{\rm tr}(\tilde{\signal{R}}_{\kappa}\tilde{\signal{C}}_n),
\end{equation}
where $\tilde{\signal{C}}_n:=E(\tilde{\signal{v}}_n\tilde{\signal{v}}^{\sf T}_n)$ is the correlation matrix of $\tilde{\signal{v}}_n$
and $J_{{\rm min}}:=E(d^2_n)-\tilde{\signal{p}}^{\sf T}\tilde{\signal{R}}^{-1}_{\kappa}\tilde{\signal{p}}$ is the minimum MSE.
We assume $e^*_n:=d_n-\tilde{\signal{\kappa}}^{\sf T}_n\tilde{\signal{\alpha}}^*$ is sufficiently close to the optimal solution of the infinite order model so that $E(e^*_n)\approx 0$, and $e^*_n$ and $\tilde{\signal{\kappa}}_n\tilde{\signal{\kappa}}^{\sf T}_n$ are uncorrelated.
Following the arguments in \cite[Section
I\hspace{-.1em}I\hspace{-.1em}I.~D]{analysisKNLMS} with
$\signal{\kappa}_{\omega}$ and $\signal{v}$ replaced respectively by
$\tilde{\signal{\kappa}}$ and $\tilde{\signal{v}}$, we arrive, with
simple manipulations, at the following recursion:\vspace{-.15cm}
\begin{align}
\label{eq:ctilde}\hspace{-.75cm}
\tilde{\signal{C}}_{n+1}\approx \tilde{\signal{C}}_n+ \eta^2(\tilde{\signal{T}}_n+J_{{\rm min}}\tilde{\signal{R}}_{\kappa})-\eta(\tilde{\signal{R}}_{\kappa}\tilde{\signal{C}}_n+\tilde{\signal{C}}_n\tilde{\signal{R}}_{\kappa}),
\end{align}
where
$\tilde{\signal{T}}_n
:=E(\tilde{\signal{\kappa}}_n\tilde{\signal{\kappa}}^{\sf T}_n\tilde{\signal{v}}_n\tilde{\signal{v}}^{\sf
T}_n\tilde{\signal{\kappa}}_n\tilde{\signal{\kappa}}^{\sf T}_n)$
and its $(\ell,m)$ component
can be approximated as
\begin{equation}
  \label{eq:that}
  [\tilde{\signal{T}}_n]_{\ell,m}\approx{\rm
  tr}(\tilde{\signal{S}}_{\ell,m}\tilde{\signal{C}}_n),
~~~ 1\leq \ell,m\leq r.
\end{equation}
Here, the $(p,q)$ component 
($1 \leq p,q\leq r$)
of
$\tilde{\signal{S}}_{\ell,m}$ is
defined as\vspace{-.2cm}
\begin{equation}
 [\tilde{\signal{S}}_{\ell,m}]_{p,q}
:=E(\tilde{\kappa}_{n,\ell}\tilde{\kappa}_{n,m}\tilde{\kappa}_{n,p}\tilde{\kappa}_{n,q})
=
\signal{g}^{\sf T}_{\ell} \signal{H}_{m,p} \
\signal{g}_{q},
\end{equation}
where
$\tilde{\kappa}_{n,\ell}:=[\tilde{\signal{\kappa}}_n]_\ell$,
$\signal{g}_{\ell}$ ($1\leq \ell\leq r$) is the
$\ell$-th column vector of $\signal{G}^{-\frac{1}{2}}$, and
$ \signal{H}_{m,p}:=E(\signal{\kappa}_n\signal{\kappa}_n^{\sf
  T}\signal{g}_{m}\signal{g}_{p}^{\sf
  T}\signal{\kappa}_n\signal{\kappa}_n^{\sf T})$.
The approximation in \refeq{eq:that} can be developed
by following the arguments in \cite[Section 3.3]{chen2014} with
$\signal{\kappa}_{\omega}$ and $\signal{v}_{\omega}$ replaced
 by $\tilde{\signal{\kappa}}$ and $\tilde{\signal{v}}$, respectively.
Finally, the $(i,j)$ component of $\signal{H}_{m,p}$
 can be written as
\begin{equation}
  \label{eq:trace}
  [\signal{H}_{m,p}]_{i,j}=\signal{g}_{m}^{\sf
  T}\signal{S}_{i,j}\signal{g}_{p},
~~~1\leq i,j\leq r,
\end{equation}
where
$[\signal{S}_{i,j}]_{s,t}:=E(\kappa_{n,i}\kappa_{n,j}\kappa_{n,s}\kappa_{n,t})$,
$1 \leq s,t\leq r$,
with $\kappa_{n,i}:=\kappa(\signal{u}_n,\signal{u}_{j_i})$
can be computed by \cite[Eq.~(35)]{chen2014}.

Let us now establish the mean-square stability condition and
derive the steady-state MSE.
Due to the presence of
$\tilde{\signal{R}}_{\kappa}\tilde{\signal{C}}_n+\tilde{\signal{C}}_n\tilde{\signal{R}}_{\kappa}$
in \refeq{eq:ctilde},
we exploit the lexicographic representation of $\tilde{\signal{C}}_n$, i.e, the
columns of each matrix are stacked on top of each other into a vector.
The recursion \refeq{eq:ctilde} can be rewritten as
\begin{equation}
\tilde{\signal{c}}_{n+1}=\signal{K}\tilde{\signal{c}}_{n}+\eta^2
J_{{\rm min}}\tilde{\signal{r}}_{\kappa},
\label{eq:rexi}
\end{equation} 
where $\tilde{\signal{c}}_{n}$ and $\tilde{\signal{r}}_{\kappa}$ are the
lexicographic forms of $\tilde{\signal{C}}_n$ and
$\tilde{\signal{R}}_{\kappa}$, respectively, and
\begin{equation}
  \label{eq:K}
  \signal{K}:=\signal{I}_{r^2} -\eta(\signal{K}_{1} + \signal{K}_{2}) + \eta^2 \signal{K}_3,
\end{equation}
where $\signal{K}_{1}:=\signal{I}_{r}\otimes
\tilde{\signal{R}}_{\kappa}$,
$\signal{K}_{2}:=\tilde{\signal{R}}_{\kappa}\otimes \signal{I}_{r}$, and
$\signal{K}_3$ is 
an $r^2\times r^2$ matrix entries are:
$[\signal{K}_3]_{\ell+(m-1)r,p+(q-1)r}:=[\tilde{\signal{S}}_{\ell,m}]_{p,q}$with
$1 \leq \ell,m,p,q\leq r$.
Here, $\otimes$ denotes the Kronecker product.
By \refeq{eq:rexi} and \refeq{eq:K}, we obtain the following results.
\begin{theorem}[Mean-square stability]
Assume CMIA holds. For any initial conditions and $\eta$ satisfying
 \refeq{eq:theorem1}, given a dictionary
 $\{\kappa(\cdot,\signal{u}_{j})\}_{j\in\mathcal{J}}$, the Natural KLMS
 algorithm with Gaussian kernel is mean-square stable, if the matrix
 $\signal{K}$ is stable (i.e., the spectral radius of $\signal{K}$ is
 less than one). 
\end{theorem}
{\it Proof:} The algorithm is said to be mean-square stable if, and only if, the state vector remains bounded and tends to a steady-state value, regardless of the initial condition \cite{sayed}.
To complete the proof, it is sufficient to show that
$\norm{\tilde{\signal{v}}_n}^2_{\signal{F}}$ remains bounded and tends to a
steady-state value, where $\signal{F}$ is a diagonal positive definite matrix.
This is verified by the fact that $\tilde{\signal{c}}_n$ is bounded and tends to a steady-state value if the matrix $\signal{K}$ is stable.
\migip

\begin{theorem}[MSE in the steady state]
Consider a sufficiently small step size $\eta$, which ensures mean and mean-square stability. The steady-state MSE is given by \refeq{eq:mse} with the lexicographic representation of $\tilde{\signal{C}}_{\infty}$ given by
\begin{equation}
 \label{eq:theorem3}
 \tilde{\signal{c}}_{\infty}= \eta^2 J_{{\rm
 min}}(\signal{I}_{r^2}-\signal{K})^{-1}\tilde{\signal{r}}_{\kappa},
\end{equation}
provided that $\signal{I}_{r^2}-\signal{K}$ is invertible.
\end{theorem}
{\it Proof:} Letting $\tilde{\signal{c}}_{n+1}=\tilde{\signal{c}}_n$
in \refeq{eq:rexi} and rearranging the equation, we obtain \refeq{eq:theorem3}.
\migip

We remark on Theorem 3 that 
the invertibility of $\signal{I}_{r^2}-\signal{K}$ is actually
ensured by the stability of $\signal{K}$.

\section{Simulation results}
\label{sec:numerical}
We shall compare simulated learning curves and analytic models to
validate the analysis.
We conduct two experiments under the same settings as in \cite{chen2014}.
In the first experiment, the input sequence is generated by
\begin{equation}
\label{eq:exinput}
u_n := \rho u_{n-1}+\sigma_u\sqrt{1-\rho^2}\omega_n,
\end{equation}
where $\omega_n$ is the noise following the i.i.d standard normal distribution.
The nonlinear system is defined as follows:
\begin{equation}
  \label{eq:2}
  \begin{cases}
x_n&:=0.5u_n-0.3u_{n-1}\\
d_n&:=x_n-0.5x_n^2+0.1x^3_n+\nu_n,
\end{cases}
\end{equation}
where $\nu_n$ is an additive zero-mean Gaussian noise with the standard deviation $\sigma_\nu=0.05$.
The input vector is $\signal{u}_n=[u_n\ u_{n-1}]^{\sf T}$. 
The step size, the standard deviation of the input, the input correlation parameter, the kernel parameter and the dictionary size are set to $\eta=0.075$, $\sigma_u=0.5$, $\rho=0.5$, $\sigma=0.7$ and $r=25$, respectively.
The dictionary is $r$ samples on a uniform grid defined on $[-1,1]\times [-1,1]$.
\begin{figure}[t]
  \centering
\psfrag{mse}{MSE}
\psfrag{iteration}{\hspace{-.5cm}Iteration number}
\psfrag{steady}{Theoretical steady state MSE}
\psfrag{learning}{Simulated curve}
\psfrag{theory}{Theoretical transient MSE}
\psfrag{parath}{Parameter (theory)}
\vspace{-.25cm}
\subfigure[]{\includegraphics[width=6.9cm]{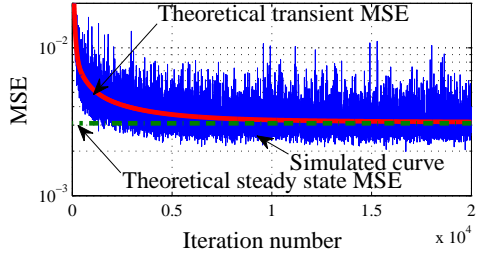}\label{fig:res1-1}}
\psfrag{mse}{MSE}
\psfrag{iteration}{\hspace{-.5cm}Iteration number}
\psfrag{select}{Selective update}
\psfrag{selective}{Selective update}
\psfrag{non}{\hspace{-1cm}Full update}
\psfrag{theory}{Theoretical transient MSE}
\psfrag{parath}{Parameter (theory)}
\subfigure[]{\includegraphics[width=6.9cm]{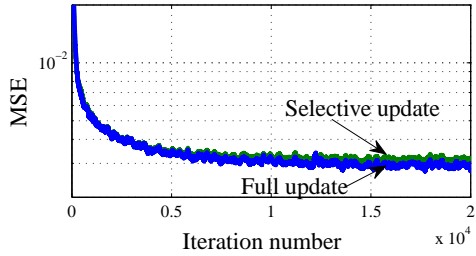}\label{fig:res1-2}}
\caption{Simulation results of the first experiment.}
\end{figure}
\begin{figure}
\centering
\psfrag{mse}{MSE}
\psfrag{iteration}{\hspace{-.5cm}Iteration number}
\psfrag{steady}{Theoretical steady state MSE}
\psfrag{learning}{Simulated curve}
\psfrag{select}{Selective update}
\psfrag{selective}{Selective update}
\psfrag{non}{Full update}

\psfrag{theory}{Theoretical transient MSE}
\psfrag{parath}{Parameter (theory)}\vspace{-.25cm}
\subfigure[]{\includegraphics[width=7.25cm]{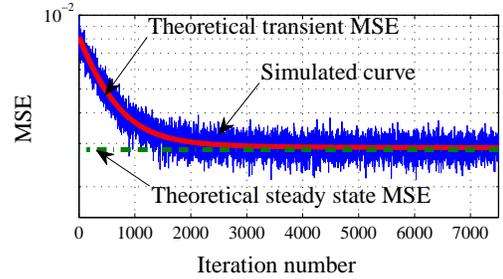} \label{fig:res2-1}}
\subfigure[]{\includegraphics[width=7.25cm]{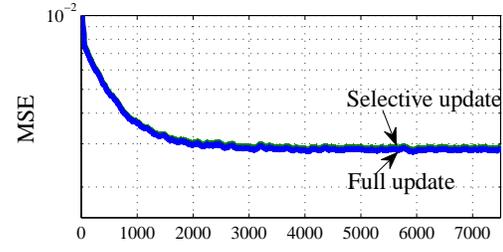}\label{fig:res2-2}}
\caption{Simulation results of the second experiment.}
\label{fig:res}
\end{figure}
\begin{table}[t]
\caption{Computational complexity of the Natural KLMS algorithm.}
\label{tb:complexity}
\vspace*{-.5em} 
\begin{center}
\begin{tabular}{|c|c|} 
\hline
Selective update&$(L+s_n+1)r+O(s_n^3)$ \\ \hline
Full update & $(L+r+2)r$\\ \hline
\end{tabular}\vspace*{-1.5em} 
\end{center}
\end{table}

Fig.~\ref{fig:res1-1} depicts the results: 
the learning curves, the theoretical transient MSE curve, and the theoretical steady state
MSE line are presented in blue, red, and green (dotted line), respectively.
The simulated curve is obtained by averaging over 300 Monte-Carlo runs. 
The theoretical MSE is estimated by \refeq{eq:mse} with $\tilde{\signal{C}}_n$ recursively evaluated by \refeq{eq:ctilde}.
The steady state MSE is computed by Theorem 3.
Although the input is correlated, the theoretical MSE presented in this paper well represents the behavior of the Natural KLMS algorithm. 

In the second experiment, the fluid-flow control problem is considered \cite{AlDuwaish1996}:
\begin{equation}
  \begin{cases}
x_n:=&0.1044u_n+0.0883u_{n-1}\\ &+1.4138x_{n-1}-0.6065x_{n-2}\\
d_n:=&0.3163x_n/ \sqrt{0.1+0.9x_n^2}+\nu_n,
\end{cases}
\end{equation}

\begin{figure}[t]
  \centering
\psfrag{comp}{\hspace{-.5cm}Complexity}
\psfrag{rn}{\hspace{-1cm}Dictionary size $r$}
\psfrag{full}{Full update}
\psfrag{hypass}{Selective update}
\psfrag{theory}{Theoretical transient MSE}
\psfrag{parath}{Parameter (theory)}
  \includegraphics[width=6.5cm]{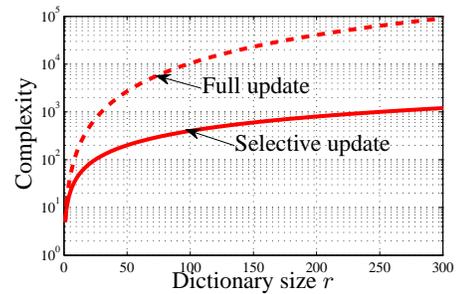}
\caption{Computational complexity.}
\label{fig:comp}
\end{figure}

\noindent
where the input $u_n$ is generated again by \refeq{eq:exinput} with $\sigma_u=0.5$ and $\rho=0.5$, and the standard deviation of the additive Gaussian noise $\nu_n$ is set to $\sigma_\nu=0.05$.
The kernel parameter is set to $\sigma=0.75$.
The input vector is $\signal{u}_n=[u_n\ u_{n-1}]^{\sf T}$.
31 dictionary elements are selected from the inputs $\signal{u}_n$ based
on the coherence criterion \cite{richard09} in advance.
The step size is set to $\eta=0.01$.
The simulated curves are obtained by averaging over 300 Monte-Carlo runs, and
the same theoretical model as the first experiment is used.
Fig \ref{fig:res2-1} depicts the results.
Again, the simulation results  show the validity of the analysis.
Table \ref{tb:complexity} summarizes the overall per-iteration complexity
(the number of real multiplications) of the Natural KLMS algorithm with
full update and selective update (see \cite{hypass,takizawa_tsp_2014}), 
and Fig.~\ref{fig:comp} illustrates the complexity as a function of the
dictionary size $r$ for $L=2$ and $s_n=1$;
$O(s_n^3)$ is counted simply as $s_n^3$.
Here, $s_n=1$ means that only one coefficient is updated at each
iteration and hence the complexity is reduced drastically.
Fig \ref{fig:res1-2} and \ref{fig:res2-2} depict the MSE learning curves of
the Natural KLMS algorithm with full update and selective update for $s_n=1$.
It can be seen that 
the Natural KLMS algorithm with the selective update
exhibits a steady-state MSE comparable to the full-update case
with drastically lower complexity.

\vspace{-.25cm}
\section{Conclusion}
\vspace{-.25cm}
\label{sec:conclusion}
This paper presented a stochastic behavior analysis of 
the Natural KLMS algorithm which is 
a stochastic restricted-gradient descent method.
The analysis provided a transient and steady-state MSEs of the algorithm.
We also derived stability conditions in the mean and mean-square sense.
Simulation results showed that the theoretical MSE curves given by the analysis 
well meet the simulated MSE curves.
The outcomes of this study will serve as a theoretical basis to
compare the performances of KNLMS and Natural KLMS.

\bibliographystyle{IEEEtran}
\bibliography{vqmetric}

\end{document}